%% file: main.tex
\documentclass[10pt,twocolumn,letterpaper]{article}

\usepackage[dvipsnames, svgnames, x11names]{xcolor}
\usepackage[pagenumbers]{cvpr} 
\usepackage{natbib}
\usepackage{csquotes}
\usepackage{booktabs}
\usepackage{xcolor}
\definecolor{lightgrey}{RGB}{240, 240, 240}
\usepackage{diagbox}
\usepackage{bbding}
\usepackage{multirow}
\usepackage{colortbl}
\usepackage{array}
\usepackage{soul}

\usepackage[breakable, theorems, skins]{tcolorbox}

\definecolor{cvprblue}{rgb}{0.21,0.49,0.74}
\usepackage[pagebackref,breaklinks,colorlinks,citecolor=cvprblue]{hyperref}
\def\paperID{8051} 
\def\confName{CVPR}
\def\confYear{2024}

\title{\includegraphics[width=0.03\textwidth]{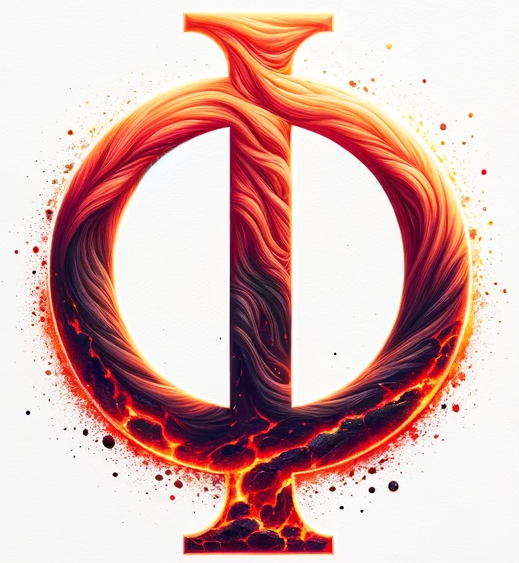} LLaVA-Phi: Efficient Multi-Modal Assistant with Small Language Model}

\author{
Yichen Zhu$^{1}$,
Minjie Zhu$^{1, 2}$,
Ning Liu$^{1}$,
Zhicai Ou$^{1}$,
Xiaofeng Mou$^{1}$,
Jian Tang$^{1}$
\\
{$\textsuperscript{1}\text{Midea Group}$,
$\textsuperscript{2}\text{East China Normal University}$
} 
}

\begin{document}
\maketitle

\begin{abstract}
In this paper, we introduce LLaVA-$\phi$ (LLaVA-Phi), an efficient multi-modal assistant that harnesses the power of the recently advanced small language model, Phi-2, to facilitate multi-modal dialogues. LLaVA-Phi marks a notable advancement in the realm of compact multi-modal models. It demonstrates that even smaller language models, with as few as 2.7B parameters, can effectively engage in intricate dialogues that integrate both textual and visual elements, provided they are trained with high-quality corpora. Our model delivers commendable performance on publicly available benchmarks that encompass visual comprehension, reasoning, and knowledge-based perception. Beyond its remarkable performance in multi-modal dialogue tasks, our model opens new avenues for applications in time-sensitive environments and systems that require real-time interaction, such as embodied agents. It highlights the potential of smaller language models to achieve sophisticated levels of understanding and interaction, while maintaining greater resource efficiency. The project is available at \href{https://github.com/zhuyiche/llava-phi}{https://github.com/zhuyiche/llava-phi}.

\end{abstract}

\section{Introduction}
Large vision language models, including Flamingo~\cite{flamingo}, GPT-4V~\cite{openai2023gpt4}, and Gemini~\cite{gemini}, have exhibited remarkable proficiency in executing instructions, engaging in multi-turn dialogues, and handling image-based question-answering tasks. The progression of open-source vision language models has been significantly propelled by the rapid advancement of open-source Large Language Models like LLaMA~\cite{llama2} and Vicuna~\cite{vicuna}. These developments primarily focus on leveraging language models with a minimum of 7B parameters, integrated with a vision encoder to enhance visual comprehension. However, this approach often results in increased test time and reduced inference speed, which are less than ideal for time-sensitive or real-time interactive applications, such as autonomous driving and robotics. This leads to an important inquiry: How effectively can small vision-language assistants perform in comparison?

Gemini~\cite{gemini} has blazed a trail for multi-modal models in mobile technology. Its streamlined variant, Gemini-Nano, boasts 1.8/3.25 billion parameters and is deployable on mobile devices. However, details like the model architecture, training data, and training methodologies remain proprietary and inaccessible to the public. In the realm of small language models, there have been notable advancements: TinyGSM~\cite{tinygsm}, with 2.6 billion parameters, achieves over 80\% accuracy on the GSM8k~\cite{gsm8k} benchmark. Additionally, models such as Phi~\cite{phi-1} have demonstrated capabilities in language understanding, commonsense reasoning, and code generation, rivaling larger language models like LLaMA-2-7B. This progress underscores the significant strides being made in the efficiency and effectiveness of smaller-scale language models.

In this paper, we introduce LLaVA-Phi, a compact vision-language assistant powered by a small language model. Our work combines the powerful open-sourced multi-modal model, LLaVA-1.5~\cite{llava1.5}, with the best-performing open-sourced small language models, Phi-2~\cite{phi1.5}. We follow a two-stage training pipeline and leverage high-quality visual instruction tuning data from LLaVA. LLaVA-Phi was evaluated across eight diverse benchmarks. Despite possessing only 3 billion parameters, it achieves performance comparable to, or even surpassing, some larger multi-modal models that are three times larger. Notably, LLaVA-Phi-3B demonstrates exceptional proficiency in ScienceQA~\cite{scienceqa}, outperforming existing large multi-modal models. Additionally, we qualitatively demonstrate LLaVA-Phi's strong generalization ability in handling challenging questions, generating code based on instructions, and solving mathematical problems.

\section{Related Work}
The rapid advancements in Large Language Models (LLMs) have significantly propelled the development of vision-language models based on LLMs. These models, representing a departure from the capabilities of the pre-LLM era, are equipped with advanced question-answering and visual comprehension skills. This progress is enabled by using LLMs as language encoding modules. Notable research in this domain includes the LLaVA-family~\cite{llava, llava1.5, llava-rlhf, llavaplus}, the BLIP-family~\cite{blip-2, instructblip}, MiniGPT-4~\cite{minigpt4}, and others. Each has demonstrated significant advancements in managing visual-centric dialogues. However, a common limitation of these open-sourced Vision-Language Models (VLMs) is their substantial computational demands, typically ranging from 7B to 65B parameters. This requirement poses challenges for deployment on edge or mobile devices, especially in real-time applications. Gemini~\cite{gemini}, a leader in this field, has released three versions of vision-language models, including the compact Gemini-Nano with 1.8B/3.25B parameters, tailored for smartphones. However, their models and data are not open-sourced. Another initiative, MobileVLM~\cite{chu2023mobilevlm}, has developed mobileLLaMA with 2.7B parameters to facilitate smaller vision-language models. Our paper explores and demonstrates the effectiveness of integrating vision-language models with open-sourced, smaller language models, assessing their potential and efficiency in a variety of applications.

\begin{figure}[t]
\centerline{
\includegraphics[width=\linewidth]{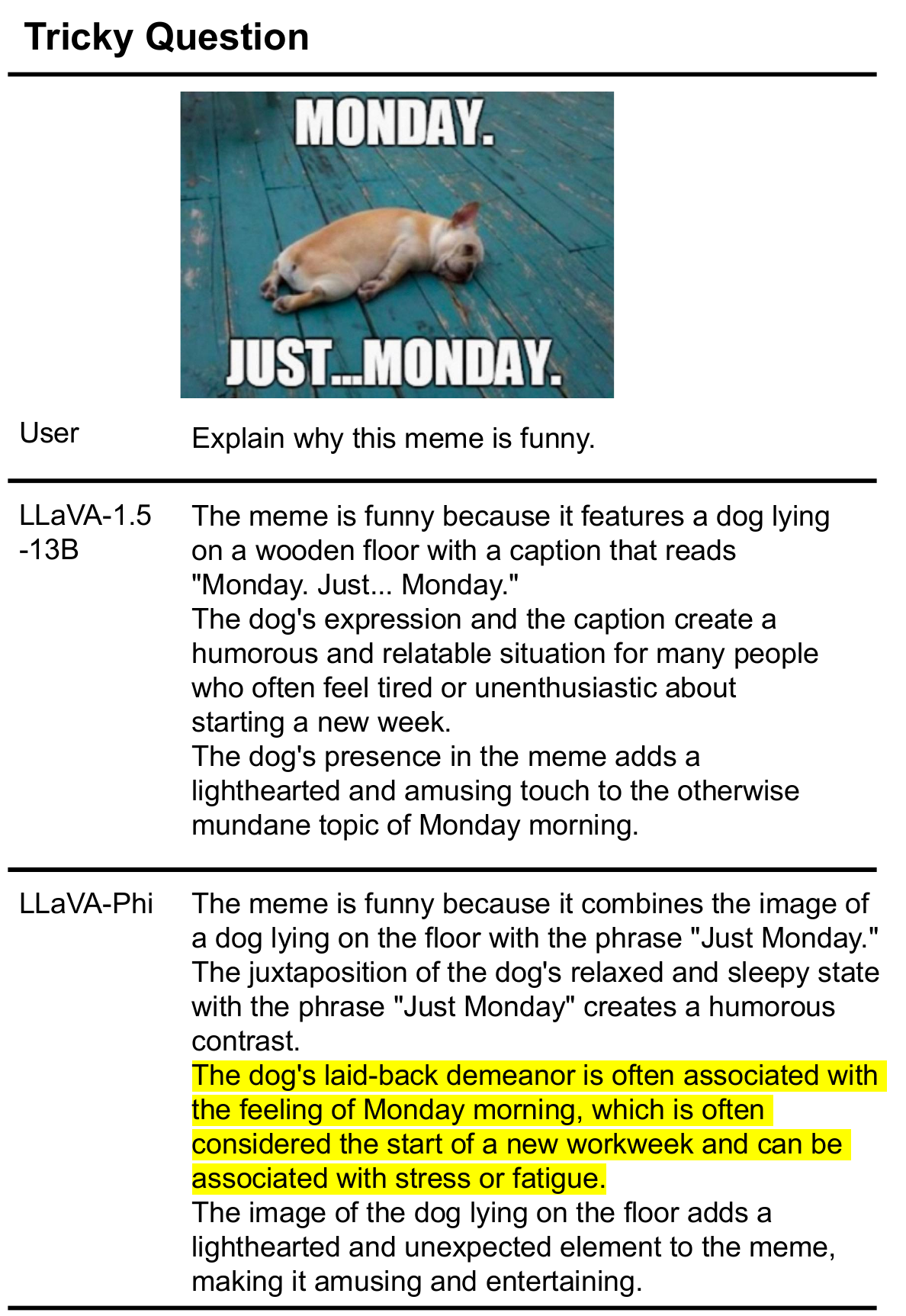}}
\caption{LLaVA-Phi is adept at identifying and responding to complex questions with empathetic reasoning.}
\label{fig:tricky}
\end{figure}

\begin{figure}[t]
\centerline{
\includegraphics[width=\linewidth]{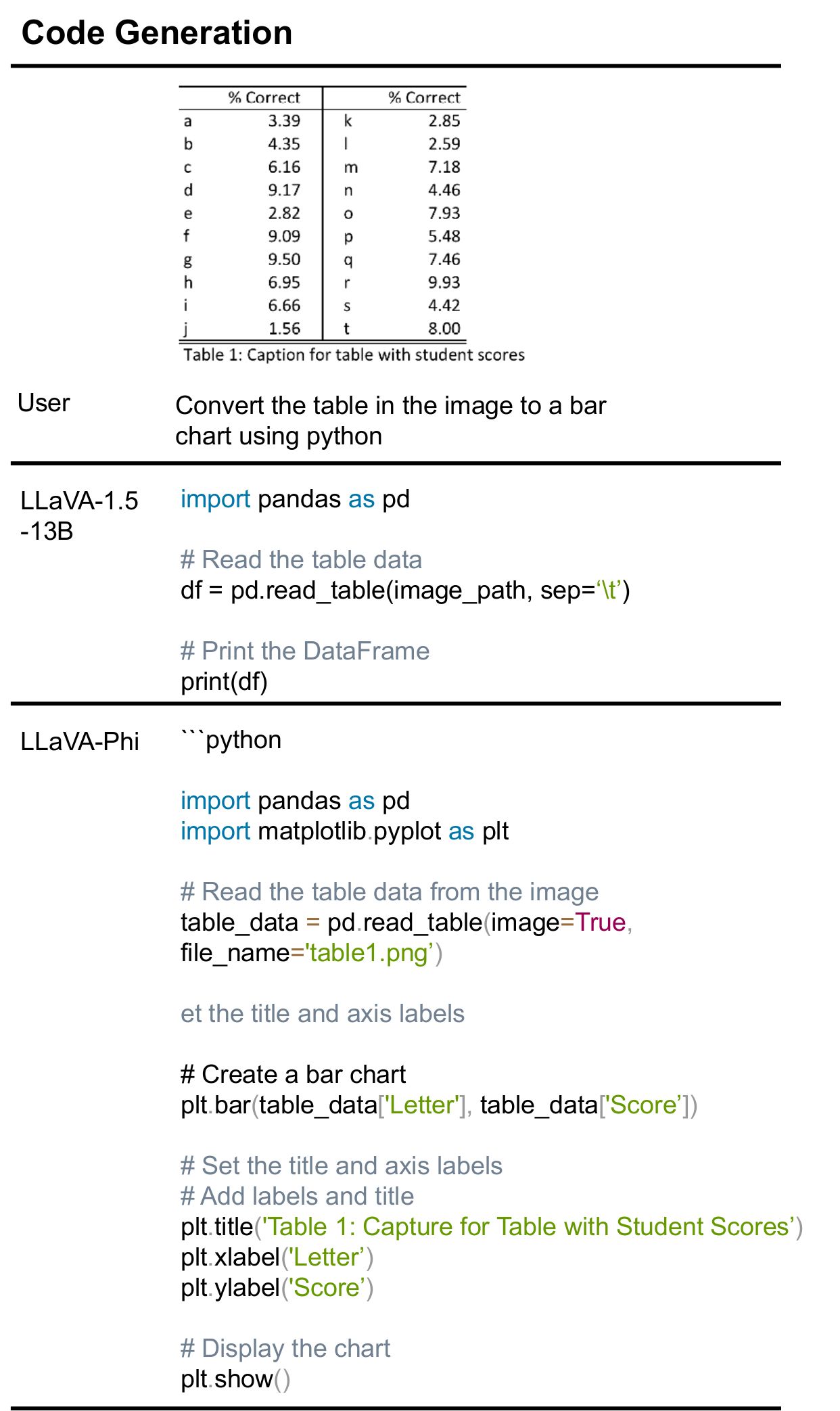}}
\caption{LLaVA-Phi can generate useful codes based on visual input and commands.}
\label{fig:code}
\end{figure}

\begin{figure}[t]
\centerline{
\includegraphics[width=\linewidth]{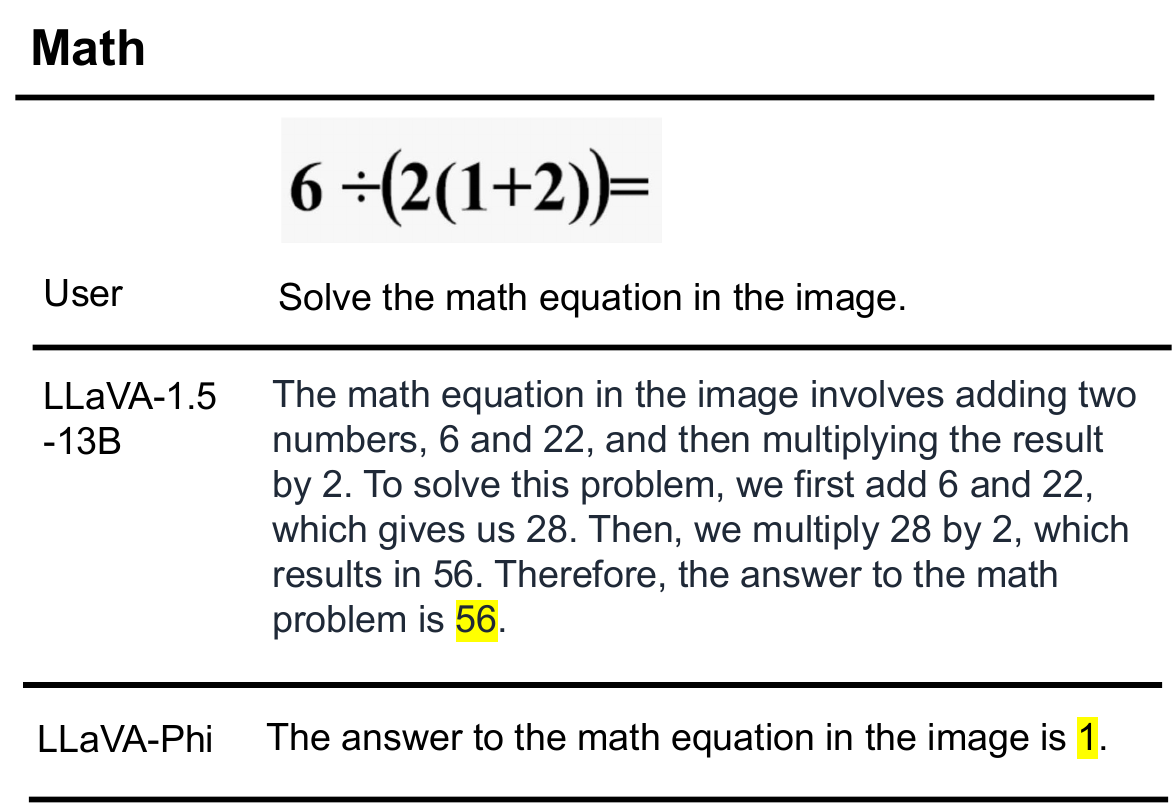}}
\caption{LLaVA-Phi is capable of performing accurate OCR on mathematical equations and solving them correspondingly..}
\label{fig:math}
\end{figure}

\section{LLaVA-Phi}
Our overall network architecture is similar to LLaVA-1.5. We use the pre-trained CLIP ViT-L/14 with a resolution of 336x336 as the visual encoder. A two-layer MLP is adopted to improve the connection of the visual encoder and LLM.
\input{main_table}

\subsection{Training}
\textbf{Supervised fine-tuning on Phi-2.} The publicly released Phi-2 model has not undergone fine-tuning. Previous research indicates that even a small amount of high-quality data can significantly enhance performance in areas such as mathematics, language reasoning, and coding tasks. In light of this, we employed supervised fine-tuning to further train Phi-2 using a select set of premium data. This data was organized in the Vicuna format. For our Supervised Fine-Tuning (SFT) data, we utilized ShareGPT from an open-source platform. The training was conducted over two epochs, beginning with an initial learning rate of 3e-5, which was linearly decreased over time. Our findings suggest that while this step might be optional, applying SFT to Phi-2 does result in modest improvements across most benchmarks. 
\\
\\
\noindent
\textbf{Training LLaVA-Phi}. 
Our training approach follows the pipeline used for LLaVA1.5, consisting of a pre-training stage and a subsequent instruction tuning phase. Initially, we kept the vision encoder and Phi-2 static, focusing exclusively on training the efficient projector. This step is followed by a comprehensive fine-tuning of both the projector and the language model (LLM), aiming to enhance their capabilities in visual comprehension and language processing.

For pre-training, we utilize a filtered subset of the CC-595K dataset~\cite{llava1.5} over one epoch, applying an initial learning rate of 1e-3 and a batch size of 256. Then, we finetune the model on LLaVA-Instruct-150K dataset for 1 epoch at a learning rate of 2e-5 and a batch size of 256. We implement a weight decay of 0.1 and utilize the Adam optimizer, characterized by momentum parameters of 0.9 and 0.98, and an epsilon value of 1e-7. We fine-tune all parameters in LLM instead of using LoRA. 
\\
\\
\textbf{Computational Cost.} Similar to LLaVA1.5, our training process is structured in two stages. For LLaVA-Phi, the pretraining phase takes 1.5 hours, followed by 8 hours dedicated to visual instruction tuning, utilizing 8 A100 GPUs. The integration of techniques such as LoRA~\cite{lora} and QLoRA~\cite{qlora} has the potential to significantly reduce training time, a possibility we plan to explore in future work.

\subsection{Qualitative Results}
We present several examples that demonstrate the remarkable generalization capabilities of LLaVA-Phi, comparing its outputs with those of the LLaVA-1.5-13B models. In Figure~\ref{fig:tricky}, a meme is displayed, and we ask the vision-language assistant to explain why this meme is considered humorous. While LLaVA-1.5-13B provides a reasonable interpretation based on the image, LLaVA-Phi's response is more empathetic, highlighting the humor by associating the dog's 'laid-back demeanor' with the 'stress or fatigue' typically associated with a 'new workweek'.

In the second example, we instructed the model to generate Python code for converting an Excel table into a bar chart, as illustrated in Figure~\ref{fig:code}. LLaVA-1.5-13B generated a simplistic code snippet that only reads the table and prints it, diverging from the instructions to create a plot. In contrast, LLaVA-Phi accurately comprehended the task, providing instructions to read the table, add a title and labels, and correctly plot the bar chart using matplotlib. We believe this enhanced code generation capability stems from Phi-2, which was pre-trained on a large corpus of code snippets and is primarily used for code generation.

The third challenge involves solving a simple math problem, requiring the model to accurately recognize text through OCR and then perform the necessary mathematical computations, as shown in Figure~\ref{fig:math}. LLaVA-1.5-13B, while providing a step-by-step computation based on the image, incorrectly recognized the numbers and mathematical symbols. In contrast, our proposed LLaVA-Phi, without providing a chain-of-thought reasoning, still produces the correct answer. Our quantitative results on ScienceQA further confirm that LLaVA-Phi excels in these types of question-answering tasks.

\section{Experiments}
We rigorously evaluated LLaVA-Phi using an extensive array of academic benchmarks specifically designed for multi-modal models. These included tests for general question-answering such as VQA-v2~\cite{vqav2}, VizWizQA~\cite{vizwiz}, ScienceQA~\cite{scienceqa}, and TextQA~\cite{textvqa}, as well as more specialized assessments like POPE~\cite{pope} for evaluating object hallucination, and MME~\cite{mme}, MMBench~\cite{mmbench}, and MMVet~\cite{mmvet} for a comprehensive evaluation of diverse multi-modal abilities, such as visual understanding and visual commonsense reasoning.

These benchmarks are meticulously structured to challenge and scrutinize complex multi-modal tasks. We benchmarked LLaVA-Phi against a variety of state-of-the-art, large vision-language models, as detailed in Table~\ref{tbl:main}. It is important to note that both our method and LLaVA1.5 utilize the same publicly available datasets for pre-training and visual instruction fine-tuning.

Our model demonstrated a capacity for visual-based question-answering, surpassing many existing large multi-modal models. Remarkably, LLaVA-Phi outperformed models that use 7B-parameter or larger Large Language Models (LLMs) as their backbone, such as IDEFICS~\cite{idefics} and InstructBLIP~\cite{instructblip}. A particularly notable achievement was our model's best performance on ScienceQA~\cite{scienceqa}. We attribute this success to the Phi-2 language model, which has been specifically trained on code generation and mathematical corpora, thereby enhancing our multi-modal model's prowess in math-based question-answering.

In the comprehensive multi-modal benchmark of MMBench~\cite{mmbench}, LLaVA-Phi showed significantly superior performance compared to many existing 7B-LLM-based vision-language models. For example, our model outperformed Otter by 11.5\% and InstructBLIP by 23.8\%. This underscores the effectiveness of LLaVA-Phi in handling complex multi-modal tasks, reinforcing the potential of smaller, more efficient models in the rapidly evolving field of multi-modal models.

We also compared to MobileVLM~\cite{chu2023mobilevlm}, a concurrent work that builds up an efficient vision-language model. Across all five benchmarks, our LLaVA-Phi consistently outperforms their method. It's important to note that the margins of lead are modest, with the exception of ScienceQA. We attribute this performance disparity primarily to the differences in the pretraining stages of the language models.

\section{Conclusion, Limitation, and Future Works}
We introduce LLaVA-Phi, a vision language assistant developed using the compact language model Phi-2. Our work demonstrates that such small vision-language models can perform effectively on standard benchmarks when combined with the LLaVA training methodology and a select dataset of high-quality data. The primary goal of our project is to aid the community in creating lightweight, multi-modal models capable of vision-language reasoning, optimized for operation on edge devices. This innovation paves the way for deploying multi-modal assistants in time-sensitive applications, such as robotics~\cite{wen2024object,zhu2024language}.
\\
\\
\noindent
\textbf{Limitations.} Given that Phi-2 utilizes the codegen-mono~\cite{codegen} tokenizer and our model has not been specifically fine-tuned for following multilingual instructions, our LLaVA-Phi architecture is unable to process instructions in multiple languages, including Chinese.
\\
\\
\noindent
\textbf{Future Works.} As language models have become significantly smaller in size compared to traditional vision-language models, they have become more accessible and affordable for the research community to explore fundamental concepts in vision-language integration. In future work, we plan to examine the impact of the size of the visual encoder and refine the training strategies for small language models, including approaches like direct preference optimization and RLHF, among other techniques. These efforts aim to further reduce model size while enhancing performance.

\newpage
{
    \small
    \bibliographystyle{ieeenat_fullname}
    \bibliography{ref}
}

\end{document}

%% file: main_table.tex
\begin{table*}[tbp]
  \centering
  \caption{Multi-modal evaluation on multi-modal benchmarks. Benchmark names are abbreviated due to space limits. $\text{VQA}^{\text{v2}}$~\cite{vqav2}; GQA~\cite{gqa}; VizWiz~\cite{vizwiz}; $\text{SQA}^{\text{I}}$: ScienceQA-IMG~\cite{scienceqa}; $\text{VQA}^{\text{T}}$: TextVQA~\cite{textvqa}; POPE~\cite{pope}; MME~\cite{mme}; MMB: MMBench~\cite{mmbench}; SEED: SEED-Bench~\cite{seed}; MM-Vet~\cite{mmvet}.}
  \label{tbl:main}
  \resizebox{1.0\linewidth}{!}{
      \begin{tabular}{l|l|cccccccc}
        \toprule
           Method & LLM  & $\text{VQA}^{\text{v2}}$ & VizWiz & $\text{SQA}^{\text{I}}$ & $\text{VQA}^{\text{T}}$ & POPE & MME & MMB  & MMVet \\
        \midrule
        Gemini-Nano2~\cite{gemini} & N/A (3.25B) & 67.5 & - & - & 65.9 & - & - & - & - \\ 
        \midrule
        OpenFlamingo~\cite{openflamingo} & MBT (7B) & - & - & - & 33.6 & - & - & 4.6 & - \\
        BLIP-2~\cite{blip-2} & Vicuna (13B) &  41.0 & 19.6 & 61.0 & 42.5 & 85.3 & 1293.8 & - & 22.4\\
        InstructBLIP~\cite{instructblip} & Vicuna (7B)  &  - & 34.5 & 60.5 & 50.1 & - & - & 36.0 & 26.2 \\
        InstructBLIP~\cite{instructblip} & Vicuna (13B) & - & 33.4 & 63.1 & 50.7 & 78.9 & 1212.8 & - & 25.6 \\
        MiniGPT-4~\cite{minigpt4} & Vicuna (7B) & - & - & - & - & - & 581.7 & 23.0 & 22.1\\
        Shikra~\cite{shikra} & Vicuna (13B) &  77.4  & - & - & - & - & - &  58.8 & - \\
        Otter~\cite{otter} & LLaMA (7B) & - & - & - & - & - & 1292.3 & 48.3 & 24.6  \\
        Qwen-VL~\cite{qwen} & Qwen (7B) & \textbf{78.8} & 35.2 & 67.1 & \textbf{63.8} & - & - & 38.2 & -\\
        Qwen-VL-Chat~\cite{qwen} & Qwen (7B) & 78.2 & 38.9 & 68.2 & 61.5 & - & 1487.5 & 60.6 & - \\
        IDEFICS-9B~\cite{idefics} & LLaMA (7B)   & 50.9 & 35.5 & - & 25.9 & - & - & 48.2 & -   \\
        IDEFICS-80B~\cite{idefics} & LLaMA (65B)   & 60.0 & 36.0 & - & 30.9 & - & - & 54.5 & -   \\
        LLaMA-Adapter-v2~\cite{Llama-adapter-v2}  & LLaMA (7B) & - & - & - & - & - & 1328.4 & 39.5 & \textbf{31.4} \\ 
        LLaVA~\cite{llava} & Vicuna (7B) & - & - & - & - & - & 502.8 & 36.2 & 28.1 \\
        LLaVA-1.5~\cite{llava1.5} & Vicuna (7B)  & 78.5 & \textbf{50.0}  & 66.8 & 58.2 & \textbf{85.9} & \textbf{1510.7} & \textbf{64.3} & 30.5 \\
        MobileVLM~\cite{chu2023mobilevlm} & M-LLaMA (2.7B) &  - & - & 61.0 & 47.5 &  84.9 & 1288.9 & 59.6 & - \\
        \midrule
        \rowcolor{lightgrey}
        LLaVA-Phi & Phi-2 (2.7B) & 71.4 & 35.9 & \textbf{68.4} & 48.6 & 85.0 & 1335.1 & 59.8 &  28.9 \\
        \bottomrule
      \end{tabular}
    }
\end{table*}